\documentclass[11pt]{article}

\usepackage[T1]{fontenc}
\usepackage[utf8]{inputenc}
\IfFileExists{lmodern.sty}{\usepackage{lmodern}}{}
\usepackage[margin=1in]{geometry}
\usepackage{microtype}
\usepackage{amsmath,amssymb}
\usepackage{graphicx}
\usepackage{booktabs}
\usepackage[numbers]{natbib}
\usepackage[hidelinks]{hyperref}
\usepackage{xcolor}

\newcommand{\E}{E}
\newcommand{\none}{\varnothing}
\newcommand{\depthstar}{a^{\star}}

\newcommand{\zpRight}{53.6}
\newcommand{\zpChoice}{43.8}
\newcommand{\zpSilent}{7.8}
\newcommand{\zpMs}{42}
\newcommand{\stRight}{58.6}
\newcommand{\stChoice}{58.4}
\newcommand{\stOrder}{7.3}
\newcommand{\zpSameCpuMs}{18}
\newcommand{\sameCpuCases}{118}
\newcommand{\clefRight}{53.3}
\newcommand{\layaOrder}{13.1}
\newcommand{\llmLo}{57.0}
\newcommand{\llmHi}{60.3}
\newcommand{\zpRightConf}{66}
\newcommand{\stRightConf}{37}
\newcommand{\zpT}{0.258}
\newcommand{\zpB}{4.52}
\newcommand{\zpDepthMin}{1.17}

\title{ZeroPoint: Lightweight Energy-Based Decision Model}
\author{Javier Mar\'in\\\texttt{javier@jmarin.info}}
\date{}

\begin{document}
\maketitle

\begin{abstract}
Energy-based models (EBMs) assign a scalar energy to every joint value of the observed and predicted variables. In decision models, the prediction is chosen from a finite list of options provided with each input. Inference then reduces to comparing the energies of the options, and training lowers the energy of the correct option relative to the others. A complete decision must also account for the possibility that no option is acceptable. However, the energies learned by a standard EBM define only relative compatibility: adding the same input-dependent quantity to all energies leaves the predictions, the Gibbs distribution, and the usual training losses unchanged. The model therefore has no learned reference level against which to decide whether any option fits. We propose ZeroPoint, an EBM that makes the empty answer a state of the model with energy zero for every input. This fixed state provides an explicit reference for the energies of all listed options, whose non-negative depths measure their compatibility with the input relative to the empty answer. The options and the empty answer share one distribution, in which an option at zero depth receives no probability. The decision to select an option or to abstain is thus part of ordinary inference, trained with the negative log-likelihood loss, rather than a separate rejection rule. The energy function is a small multilingual encoder that reads the input together with all the options and is invariant to their order. Ablations show that most of the benefit comes from training with examples in which no option fits. With the same encoder and data, the fixed zero separates inputs with and without an acceptable option slightly better than a learned rejection class and clearly better than a thresholded softmax, while leaving the choice among options unchanged.
\end{abstract}

\section{Introduction}

Decision systems receive a text and a short list of options and must select one, such as the queue for a support ticket~\cite{casanueva2020efficient,larson2019evaluation} or the tool for a request~\cite{schick2023toolformer}. Often, the correct option is not on the list~\cite{larson2019evaluation}, and the useful response is then to abstain rather than return the most probable option~\cite{chow1970optimum,geifman2017selective,hendrycks2016baseline}. EBMs are a natural framework for this problem~\cite{lecun2006tutorial}. Each option is assigned an energy that is low when the option fits the text, and the model selects the option with the lowest energy. The same framework explains why abstention is difficult. Energies are only compared with each other~\cite{lecun2006tutorial,liu2020energy}. Training lowers the energy of the correct option and raises the energies of the incorrect ones, but nothing in this procedure fixes a reference level for the energy. Consequently, a list of four inadequate options and a list of four adequate ones may yield energies that differ only by an additive constant, and the model cannot distinguish one list from the other.

Energies learned only through their gradients, as in Hamiltonian world models~\cite{greydanus2019hamiltonian,yalovetzky2026lehami}, support prediction but are likewise defined only up to an additive constant. We ask what is required for a learned energy to have a meaningful zero, and what form a decision model takes once it has one. Our findings are the following.

\textbf{The missing zero is a symmetry of the training objective.} Losses built on a Gibbs distribution are invariant to adding any function of the input to all energies. Rejection by an energy threshold therefore rests on a quantity the model never learns.

\textbf{Fixing the energy of the empty answer removes this symmetry.} With the empty answer at zero energy for every input, abstention becomes a comparison with a depth learned by the same likelihood as the rest of the model. This generalizes the fixed energy of the non-face answer in the face detector of~\cite[Sec.~4.1]{lecun2006tutorial} from a binary decision to lists of options, with the level learned rather than fixed in advance.

\textbf{Training with examples in which no option fits matters more than the form of the distribution.} With the same encoder and data, the fixed zero and a learned rejection class reach the same rate of correct decisions, while a softmax trained without such examples falls far behind. The fixed zero adds a small but consistent improvement in separating inputs with and without a correct option, and it does not change the choice among options.

\textbf{The resulting model trades ranking accuracy for efficiency.} ZeroPoint ranks options less accurately than larger decision models, but it runs on a CPU, is considerably smaller, and is exactly invariant to the order of the options.

\section{The reference level of the energy}
\label{sec:gauge}

\subsection{Energy-based decisions}

A model with parameters $W$ assigns a scalar energy $\E(W,Y,X)$ to every input $X$ and answer $Y$ from a set $\mathcal{Y}$, with low energies corresponding to compatible pairs. Inference is energy minimization, $Y^{*}=\arg\min_{Y\in\mathcal{Y}}\E(W,Y,X)$~\cite{lecun2006tutorial}. To read energies as probabilities, we use the Gibbs distribution~\cite{gibbs1902} at inverse temperature $\beta$
\begin{equation}
P(Y\mid X)=\frac{e^{-\beta \E(W,Y,X)}}{\sum_{y\in\mathcal{Y}}e^{-\beta \E(W,y,X)}},
\qquad
F_\beta(X)=-\frac{1}{\beta}\log\sum_{y\in\mathcal{Y}}e^{-\beta \E(W,y,X)} .
\label{eq:gibbs}
\end{equation}
The denominator is a sum over all answers in $\mathcal{Y}$, and $F_\beta$ is the free energy of the input, a soft minimum of its energies. Training minimizes a loss functional averaged over examples $(X^i,Y^i)$. The negative log-likelihood loss is
\begin{equation}
L_{\mathrm{nll}}(W,Y^i,X^i)=\E(W,Y^i,X^i)-F_\beta(X^i).
\label{eq:nll}
\end{equation}
Its gradient decreases the energy of the correct answer and increases the energy of each incorrect answer in proportion to its probability~\cite{lecun2006tutorial}. The hinge loss acts only on the most offending incorrect answer $\bar Y^i$, defined as the incorrect answer of the lowest energy, and enforces a margin $m$ between its energy and that of the correct answer,
\begin{equation*}
  \max\bigl(0,\; m+\E(W,Y^i,X^i)-\E(W,\bar Y^i,X^i)\bigr).
\end{equation*}

\subsection{The additive freedom and rejection}

Suppose that a function of the input alone, $c(X)$, is added to every energy, so that $\E'(W,Y,X)=\E(W,Y,X)+c(X)$. The factor $e^{-\beta c(X)}$ then appears in both the numerator and the denominator of \eqref{eq:gibbs} and cancels, and the minimizer $Y^{*}$ is unchanged. Both losses above depend only on differences between energies at the same input, so they are unchanged as well. Only the free energy shifts, $F'_\beta(X)=F_\beta(X)+c(X)$. The data therefore determine the energy of each answer only relative to the other answers for the same input. The offset $c(X)$ is the gauge freedom of the model, in the sense used in physics. It leaves every prediction and every gradient unchanged, and training does not constrain it.

This invariance does not affect the choice of the best answer, which depends only on differences between energies. It does affect the decision of whether any answer is acceptable, which requires comparing an energy with a fixed level. The standard energy-based formulation of this decision compares the lowest energy with a threshold and rejects when $\min_y \E(W,y,X)>\tau$~\cite[Sec.~1.2]{lecun2006tutorial}. Under the transformation $\E\mapsto\E+c(X)$, the test becomes $\min_y\E(W,y,X)+c(X)>\tau$, and its outcome depends on a function that no term of the loss constrains. Two alternatives place the threshold on other quantities. The first thresholds the probability of the most probable answer~\cite{chow1970optimum,hendrycks2016baseline,geifman2017selective}. This probability is invariant to $c(X)$, but \eqref{eq:gibbs} normalizes over the listed answers only, so the probability mass is necessarily distributed among them. When all options are equally inadequate, each receives $1/|\mathcal{Y}|$, exactly as when all are equally adequate, and an option only slightly less inadequate than the others receives most of the mass, since probabilities grow exponentially with energy differences. The second thresholds the free energy~\cite{liu2020energy}, which is precisely the quantity shifted by $c(X)$. Its success therefore depends on regularities that training happens to produce, not on a constraint imposed by the loss. In both cases, abstention is decided outside the model.

\section{ZeroPoint}
\label{sec:model}

\subsection{The empty answer as a state at zero}

ZeroPoint receives a text $X$ and a list of options $\mathcal{Y}=\{y_1,\dots,y_n\}$. Its answer set is $\mathcal{Y}\cup\{\none\}$, where the empty answer $\none$ means that no option fits. The empty answer is not given to the encoder and has no representation of its own. For each option, the encoder computes a depth $a_y\ge 0$, which depends on the text in the list, and the energy of the option is minus that depth. The energy of the empty answer is zero for every input
\begin{equation}
\E(W,y,X)=-a_y(X,\mathcal{Y})\le 0,\qquad \E(W,\none,X)=0 .
\label{eq:energies}
\end{equation}
Adding a function $c(X)$ to every energy would move the energy of $\none$ away from zero, which \eqref{eq:energies} does not allow. The additive freedom of Section~\ref{sec:gauge} is therefore removed, and every depth is measured from the energy of the empty answer.

Fixing the energy of one answer has a precedent. In the energy-based face detector of~\cite[Sec.~4.1]{lecun2006tutorial}, the answer ``non-face'' has a fixed energy $\theta$,
\begin{equation*}
  \E(Y,X)=Y\,G_{\mathrm{face}}(X)+(1-Y)\,\theta,
\end{equation*}
and the detector reports a face when $G_{\mathrm{face}}(X)<\theta$. The decision is again a comparison with a level that does not depend on $X$. ZeroPoint differs from this construction in three respects. The empty answer competes with a list of options that changes from one input to the next, not with a single alternative. Its weight in the distribution carries a parameter learned by the negative log-likelihood, whereas $\theta$ is a fixed constant of the detector. Finally, the weight of each option is measured relative to the zero level, so that an option at that level receives no probability.

The probabilities are obtained by normalizing non-negative weights over the $n+1$ states, with a temperature $T=1/\beta$
\begin{equation}
w_y=e^{a_y/T}-1=e^{-\beta \E(W,y,X)}-e^{-\beta \E(W,\none,X)},\qquad
w_\none=e^{b},\qquad
P(y\mid X)=\frac{w_y}{w_\none+\sum_{j}w_{y_j}} ,
\label{eq:weights}
\end{equation}
and $P(\none\mid X)$ is $w_\none$ divided by the same sum. The weight of an option is its Boltzmann factor minus the Boltzmann factor of the zero level. The distribution is therefore a Gibbs distribution over the effective energies $\tilde\E(y)=-T\log\bigl(e^{a_y/T}-1\bigr)$ and $\tilde\E(\none)=-Tb$, not over the energies of \eqref{eq:energies}. Two features of \eqref{eq:weights} carry the argument. An option of zero depth has zero weight, so it has the energy of the empty answer but receives no probability. The constant $b$ sets the weight of the empty answer and can be read as the logarithm of a multiplicity, since the empty answer stands for every answer that is not on the list. Both $T$ and $b$ are scalars shared by all inputs. They are trained with the encoder by the negative log-likelihood and then refitted by the same loss on held-out lists (Section~\ref{sec:learning}). Neither depends on $X$, so neither restores the additive freedom.

\begin{figure}[t]
\centering
\includegraphics[width=0.85\linewidth]{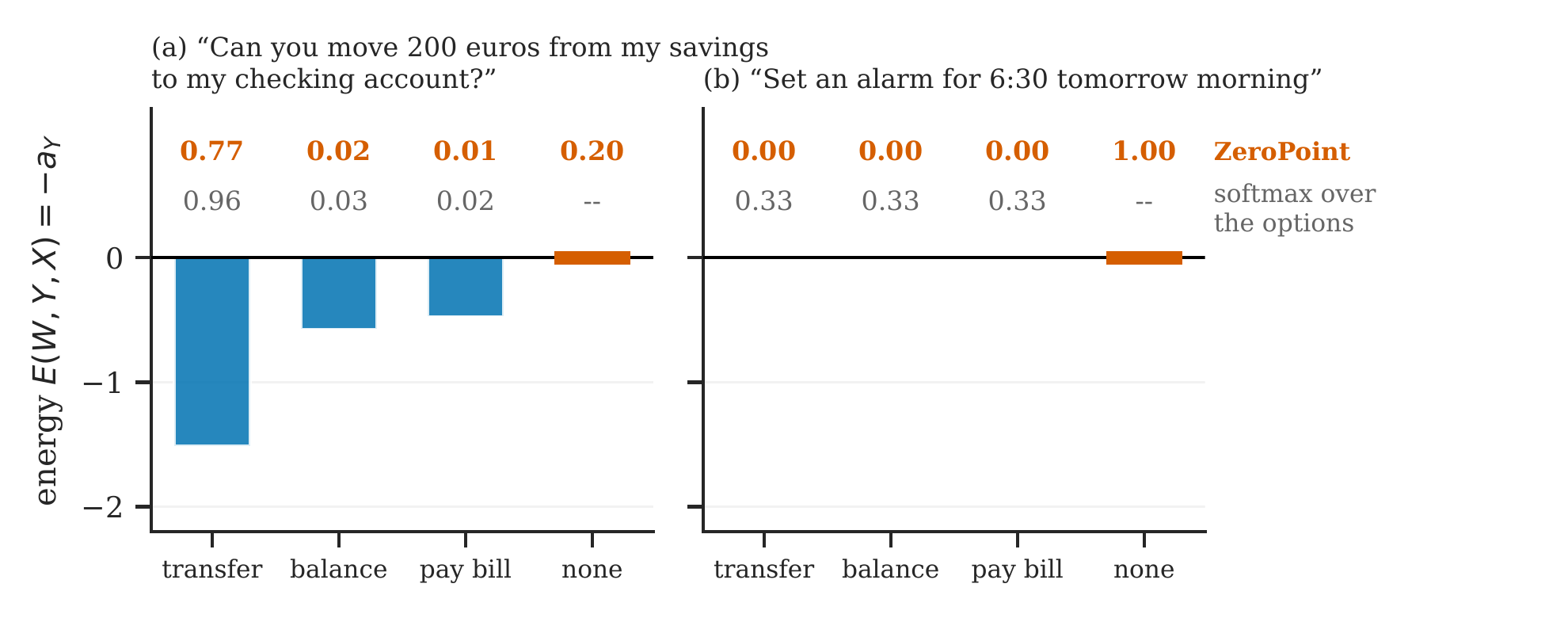}
\caption{The empty answer keeps probability away from inadequate options. Bars show the energies $\E=-a_y$ of three options for two requests, computed by the released model, and the empty answer lies at zero. Upper numbers are ZeroPoint's probabilities from \eqref{eq:weights}. Lower numbers are a softmax of $a_y/T$ over the options alone. (a) A banking request. Both rules select ``transfer'', the deepest well. ZeroPoint keeps 0.20 on the empty answer, whereas the softmax assigns 0.96 to ``transfer''. (b) An alarm request. No option has an appreciable depth. The softmax assigns about 0.33 to each of three inadequate options, and ZeroPoint assigns 1.00 to the empty answer.}
\label{fig:landscape}
\end{figure}

Three properties follow from \eqref{eq:energies} and \eqref{eq:weights}.

\begin{enumerate}
    \item \textbf{No dilution by irrelevant options.} An option of zero depth leaves every other probability unchanged, because its weight is zero. If all depths are close to zero, $w_y\approx a_y/T$, and $P(\none\mid X)=e^b/\bigl(e^b+\sum_y w_y\bigr)$ remains close to one as long as $\sum_y a_y/T$ is small compared with $e^b$. With plain Boltzmann factors $e^{a_y/T}$, each option of zero depth would have weight one, and the empty answer would receive $e^b/(e^b+n)$, which decreases as the list grows. In the released model the depth is a softplus, which is strictly positive, so an inadequate option never has exactly zero weight. On the cases without a correct option, the median depth is 0.037, or 0.14 in units of $T$, and the summed weight of the options exceeds $e^b$ in 22\% of them (Appendix~\ref{sec:length}). Figure~\ref{fig:landscape} shows two requests processed by the released model. When one option fits, both rules select it. When none fits, the softmax spreads its probability evenly over inadequate options, while ZeroPoint assigns it to the empty answer.

    \item \textbf{Abstention is a fixed depth.} The model returns the most probable state. Since $w_y$ increases with $a_y$, the empty answer is the most probable state if and only if every depth satisfies
    \begin{equation}
    a_y\le\depthstar=T\log\bigl(1+e^{b}\bigr).
    \label{eq:astar}
    \end{equation}
    The value $\depthstar$ is the same for every input and every list. In the released model, $T=\zpT$ and $b=\zpB$, so $\depthstar\approx\zpDepthMin$. A threshold therefore still exists, but it is expressed on the absolute scale defined by the zero level and obtained from the same likelihood as the rest of the model, not from a separate criterion. The depths themselves, and hence $P(\none\mid X)$, depend on the whole list, because the encoder reads all options together.

    \item \textbf{The scale remains free, the offset does not.} The distribution depends on the depths only through $a_y/T$. Multiplying all depths and $T$ by the same positive constant therefore leaves every probability unchanged, so the model retains a scale freedom but no additive freedom. This freedom does not affect any decision, and depths are meaningful only in units of temperature. In the released model, $\depthstar/T=\log\bigl(1+e^{b}\bigr)\approx 4.53$.
\end{enumerate}

\subsection{The form of the weights}
\label{sec:weights}

Fixing the energy of the empty answer does not determine how the weight of an option should depend on its depth. Three requirements single out \eqref{eq:weights}. First, an option at the zero level has the energy of the empty answer, and adding it to the list should leave every other probability unchanged, so its weight must vanish, $g(0)=0$. Second, a deeper well should receive more weight, so $g$ must increase with $a$. Third, wells that clearly fit should compare with each other as in the Gibbs distribution \eqref{eq:gibbs}, so $g(a)$ must approach $e^{a/T}$ when $a\gg T$. The function $g(a)=e^{a/T}-1$ satisfies all three. It also has a direct reading in statistical mechanics. If a well of depth $a$ is a continuum of states whose energies are uniformly distributed between $-a$ and $0$, with one state per unit of temperature, its partition function is
\begin{equation}
\frac{1}{T}\int_{0}^{a} e^{s/T}\,ds = e^{a/T}-1 .
\label{eq:well}
\end{equation}
The weight of an option is then the partition function of its well, and a well of zero depth contains no states. For shallow wells $g(a)\approx a/T$, so the weight grows linearly with depth, and for deep wells $g(a)\approx e^{a/T}$.

Two nearby alternatives each fail one requirement. The plain Boltzmann factor $g(a)=e^{a/T}$, with a constant weight $e^b$ for the empty answer, also fixes a reference level, but it gives weight one to an option of zero depth, so the probability of the empty answer falls as $e^b/(e^b+n)$ when $n$ irrelevant options are added. A learned rejection class whose weight $e^{r(X)}$ depends on the input avoids this dilution only if $r$ learns to compensate the number and depth of the options, and it restores an input-dependent reference, since adding $c(X)$ to the energies of the options and of the rejection class leaves the distribution unchanged. Appendix~\ref{sec:headablation} compares these forms with the same encoder and data.

\subsection{The training objective}

Training minimizes the negative log-likelihood \eqref{eq:nll} of the distribution \eqref{eq:weights} over the extended answer set, with $\none$ as the target when the correct option is not on the list. The fixed state changes what this loss can do. When the target is an option, the loss deepens its well and fills the others, and the empty answer counts as one more incorrect answer, so the well of the correct option must become deeper than $\depthstar$ for that option to be selected. When the target is $\none$, the gradient cannot lower the energy of $\none$, which is fixed at zero. Apart from the single scalar $b$, the only way to raise $P(\none\mid X)$ is to fill every well towards zero. The meaning of zero depth is thus learned from examples in which no option fits, which a softmax over the options alone cannot represent.

\subsection{Architecture}
\label{sec:arch}

The encoder is multilingual-E5-small~\cite{wang2024multilingual}, with 12 layers and hidden size 384, and reads one token sequence, \texttt{[CLS] query: text [SEP]} $y_1$ \texttt{[SEP]} $\dots$ $y_n$ \texttt{[SEP]}. The text receives positions $0,\dots,m-1$, and every option receives positions starting at $m$, the same for all options. Attention is bidirectional and unmasked between options. The depth of an option is the softplus of a linear function of the mean of its last hidden states. Self-attention without a mask is equivariant to permutations of tokens that share position ids~\cite{zaheer2017deep}, so permuting the options permutes their depths. Lists longer than 20 options are read in blocks of 20 after sorting the options in a canonical order, the byte order of their Unicode NFC form, which makes the floating-point computation identical for every input order. Duplicate options are rejected, since two copies of the same well would double its weight. The vocabulary was reduced to 107,299 pieces covering 25 languages, and the network is exported to ONNX~\cite{onnxruntime} with float16 weights and int8 vocabulary embeddings, 89~MB in total. Equation~\eqref{eq:weights} is evaluated in float64.

\section{Experimental evaluation}
\label{sec:exp}

This section describes the training of the released model and compares it with other decision systems. Appendices~\ref{sec:lineage} and~\ref{sec:headablation} report two ablations of the fixed zero, and Appendices~\ref{sec:eight} to~\ref{sec:repro} report further measurements of the released model.

\subsection{Training details}
\label{sec:learning}

The released model was trained on 76,211 lists for 2 epochs, in batches of 16 lists, with AdamW at a learning rate of $3\times10^{-5}$ and weight decay 0.01 for the encoder and the head, and $10^{-2}$ for $\log T$ and $b$. The learning rate warms up linearly and then decays linearly to zero. The head starts with zero weights and a bias of $\log(e-1)$, so every well starts at depth exactly 1. Most lists hold 4 to 20 options from one dataset. In 30\% of the lists the correct option is removed and the target is $\none$, in 50\% the incorrect options are those most easily confused with the correct one, the most offending incorrect answers of~\cite{lecun2006tutorial}, and 8\% are long lists of 21 to 150 options. The texts come from CLINC150 without its out-of-scope texts~\cite{larson2019evaluation}, Banking77~\cite{casanueva2020efficient}, MASSIVE in English and Spanish~\cite{fitzgerald2023massive}, four Bitext customer-service sets~\cite{bitext2023}, and 5,000 examples written for the project to cover long support tickets, negations and categories that depend on a name. Training took 24 minutes on an NVIDIA L4 GPU. After training, $T$ and $b$ alone were refitted by the same likelihood on 982 held-out lists. This refit changes the balance between acting and abstaining without changing the depths. Because of how the held-out lists were split, the 982 calibration lists contain no written examples. On the 1,281 lists used for early stopping, it reduced the negative log-likelihood from 0.452 to 0.438 and the expected calibration error from 3.7\% to 1.7\%.

\subsection{Evaluation protocol}
\label{sec:protocol}

\paragraph{Benchmark.} We use the benchmark released with the model~\cite{marin2026zeropoint}. Three tests are unlikely to be in the training data of any system. The 2,000-case test contains everyday requests and support tickets with 4 to 9 options. In the two ticket tests, with 8 and 9 options, a case without a correct option either has the ticket's own queue removed from the list or is an unrelated text, a film review~\cite{maas2011learning} or a newsgroup post~\cite{lang1995newsweeder}. Every system reads the same 300 cases per test, half of them with a correct option, 900 cases in all.

\paragraph{Systems.} We compare ZeroPoint with two open decision models, Strands Decider 2B~\cite{strands2026decider} and Clef-flash~\cite{cloudflare2026clef}, with Laya~\cite{convai2026laya}, Semantic Router~\cite{aurelio2024semanticrouter}, zero-shot NLI~\cite{yin2019benchmarking}, and six hosted language models~\cite{anthropic2026claude,openai2026gpt6}. Systems without an answer of their own for abstention receive one more option, ``None of these''. ZeroPoint reads the list as it is. Every system runs with its default settings. Systems that run on a CPU used a 2-core CPU with 2 threads, the two open decision models an NVIDIA L4 GPU, and the hosted models their APIs.

\paragraph{Measures.} A correct decision is the correct option when there is one, and abstention when there is none. Ranking is measured on the 450 cases with a correct option, as the share in which that option has the highest score. Silent mistakes are measured on the 450 cases without a correct option, as the share in which an option is selected with probability at least 0.9. For the hosted models, the probability is the confidence they state. Order changes are the share of cases whose answer changes when the options are shuffled. The measures and the threshold of 0.9 were fixed before any result was seen. Differences between systems are computed on the same cases, with 95\% intervals from 2,000 paired bootstrap resamples~\cite{efron1994introduction,koehn2004statistical}.

\subsection{Comparison with other systems}
\label{sec:comparison}

\begin{table}[t]
\centering
\small
\begin{tabular}{lrrrrrl}
\toprule
System & Correct & Ranking & Silent & Order & Latency & Runs on \\
 & decisions & & mistakes & changes & (median) & \\
\midrule
\textbf{ZeroPoint} & 53.6 & 43.8 & 7.8 & 0.0 & 42 ms & CPU \\
Strands Decider 2B & 58.6 & 58.4 & 0.0 & 7.3 & 93 ms$^{a}$ & GPU \\
Clef-flash & 53.3 & 56.2 & 5.8 & 0.0 & 156 ms & GPU \\
Laya & 49.9 & 55.6 & 4.9 & 13.1 & 593 ms & CPU \\
Semantic Router & 19.4 & 38.9 & -- & 0.6 & 17 ms & CPU \\
Zero-shot NLI & 20.7 & 41.3 & 22.2 & 0.0 & 1,697 ms & CPU \\
\midrule
Claude Haiku 5.5 & 57.0 & -- & 1.1 & -- & 778 ms & API \\
Claude Sonnet 5.5 & 59.6 & -- & 1.8 & -- & 1,520 ms & API \\
Claude Opus 5.5 & 60.3 & -- & 0.9 & -- & 2,123 ms & API \\
GPT-6 Luna & 58.0 & -- & 22.7 & -- & 1,590 ms & API \\
GPT-6.1 Terra & 58.9 & -- & 24.4 & -- & 2,010 ms & API \\
GPT-6.1 Sol & 59.9 & -- & 26.0 & -- & 2,329 ms & API \\
\bottomrule
\end{tabular}
\caption{The three tests, 900 cases per system, in percent except latency. Systems are ZeroPoint~\cite{marin2026zeropoint}, Strands Decider 2B~\cite{strands2026decider}, Clef-flash~\cite{cloudflare2026clef}, Laya~\cite{convai2026laya}, Semantic Router~\cite{aurelio2024semanticrouter}, zero-shot NLI~\cite{yin2019benchmarking} with DeBERTa-v3-base~\cite{he2021debertav3,laurer2024less}, and hosted models~\cite{anthropic2026claude,openai2026gpt6}. The hosted models return no score for every option, so ranking and order changes do not apply to them. Semantic Router returns no probability. $^{a}$On the same 2-core CPU as ZeroPoint, 6,770~ms.}
\label{tab:main}
\end{table}

\begin{table}[t]
\centering
\small
\begin{tabular}{lccc}
\toprule
System & Correct decisions & Ranking & Silent mistakes \\
\midrule
Strands Decider 2B & $-$5.0 [$-$8.6, $-$1.6] & $-$14.7 [$-$20.2, $-$9.3] & +7.8 [+5.3, +10.4] \\
Clef-flash & +0.2 [$-$3.2, +3.8] & $-$12.4 [$-$17.8, $-$7.3] & +2.0 [$-$0.9, +4.9] \\
Laya & +3.7 [+0.1, +7.3] & $-$11.8 [$-$16.4, $-$6.9] & +2.9 [$-$0.4, +6.0] \\
Semantic Router & +34.1 [+30.1, +38.0] & +4.9 [+0.2, +10.0] & -- \\
Zero-shot NLI & +32.9 [+28.7, +36.8] & +2.4 [$-$2.4, +7.8] & $-$14.4 [$-$18.7, $-$10.4] \\
\midrule
Claude Haiku 5.5 & $-$3.4 [$-$6.8, $-$0.1] & -- & +6.7 [+4.2, +9.3] \\
Claude Sonnet 5.5 & $-$6.0 [$-$9.4, $-$2.7] & -- & +6.0 [+3.3, +8.7] \\
Claude Opus 5.5 & $-$6.8 [$-$10.2, $-$3.4] & -- & +6.9 [+4.4, +9.6] \\
GPT-6 Luna & $-$4.4 [$-$7.9, $-$1.1] & -- & $-$14.9 [$-$18.9, $-$10.9] \\
GPT-6.1 Terra & $-$5.3 [$-$8.7, $-$2.0] & -- & $-$16.7 [$-$20.7, $-$12.4] \\
GPT-6.1 Sol & $-$6.3 [$-$9.7, $-$3.0] & -- & $-$18.2 [$-$22.4, $-$14.2] \\
\bottomrule
\end{tabular}
\caption{ZeroPoint minus each system, in percentage points on the same 900 cases, with 95\% paired bootstrap intervals over 2,000 resamples~\cite{efron1994introduction,koehn2004statistical}. For silent mistakes, a positive value means that ZeroPoint makes more. Systems as in Table~\ref{tab:main}.}
\label{tab:paired}
\end{table}

\begin{table}[t]
\centering
\small
\begin{tabular}{lrrrrrr}
\toprule
 & \multicolumn{3}{c}{Correct decisions} & \multicolumn{3}{c}{Silent mistakes} \\
\cmidrule(lr){2-4}\cmidrule(lr){5-7}
System & 2,000 & Removed & Unrelated & 2,000 & Removed & Unrelated \\
\midrule
\textbf{ZeroPoint} & 76.7 & 22.3 & 61.7 & 4.7 & 18.7 & 0.0 \\
Strands Decider 2B & 88.0 & 21.7 & 66.0 & 0.0 & 0.0 & 0.0 \\
Clef-flash & 85.3 & 19.3 & 55.3 & 8.0 & 9.3 & 0.0 \\
Laya & 72.3 & 19.3 & 58.0 & 14.0 & 0.7 & 0.0 \\
Semantic Router & 32.3 & 12.3 & 13.7 & -- & -- & -- \\
Zero-shot NLI & 34.3 & 15.7 & 12.0 & 20.0 & 38.0 & 8.7 \\
\midrule
Claude Haiku 5.5 & 88.7 & 16.0 & 66.3 & 1.3 & 2.0 & 0.0 \\
Claude Sonnet 5.5 & 93.7 & 17.3 & 67.7 & 0.7 & 4.7 & 0.0 \\
Claude Opus 5.5 & 92.7 & 18.7 & 69.7 & 2.0 & 0.7 & 0.0 \\
GPT-6 Luna & 91.3 & 20.3 & 62.3 & 11.3 & 52.7 & 4.0 \\
GPT-6.1 Terra & 92.7 & 19.0 & 65.0 & 8.7 & 61.3 & 3.3 \\
GPT-6.1 Sol & 93.0 & 21.0 & 65.7 & 9.3 & 66.0 & 2.7 \\
\bottomrule
\end{tabular}
\caption{Results per test, 300 cases each, in percent. ``2,000'' is the 2,000-case test, ``Removed'' the ticket test in which the cases without a correct option have their own queue removed, and ``Unrelated'' the ticket test in which they are film reviews~\cite{maas2011learning} or newsgroup posts~\cite{lang1995newsweeder}. Systems as in Table~\ref{tab:main}.}
\label{tab:pertest}
\end{table}

\paragraph{Ranking.} When a correct option exists, ZeroPoint ranks it first in \zpChoice\% of cases, against \stChoice\% for Strands Decider 2B, 56.2\% for Clef-flash and 55.6\% for Laya (Table~\ref{tab:main}). All three differences exclude zero (Table~\ref{tab:paired}). On support tickets, ZeroPoint selects the correct queue in about a quarter of cases. Ranking is the main weakness of the model.

\paragraph{Correct decisions.} ZeroPoint reaches \zpRight\% correct decisions. Strands Decider is ahead at \stRight\%, and the six hosted models reach \llmLo\% to \llmHi\%. Every one of these differences excludes zero (Table~\ref{tab:paired}). Clef-flash is level at \clefRight\% despite ranking 12.4 points better, and Laya is behind at 49.9\%. Abstention accounts for the gap between ranking and correct decisions. On the cases without a correct option, ZeroPoint selects an option at any confidence in 31.8\% of them, against 39.3\% for Strands Decider, 49.1\% for Clef-flash and 54.9\% for Laya. The gap is concentrated on the 2,000-case test, where the larger systems lead by 9 to 17 points, while on the two ticket tests ZeroPoint lies within 8 points of every decision model and hosted model (Table~\ref{tab:pertest}).

\begin{figure}[t]
\centering
\includegraphics[width=0.9\linewidth]{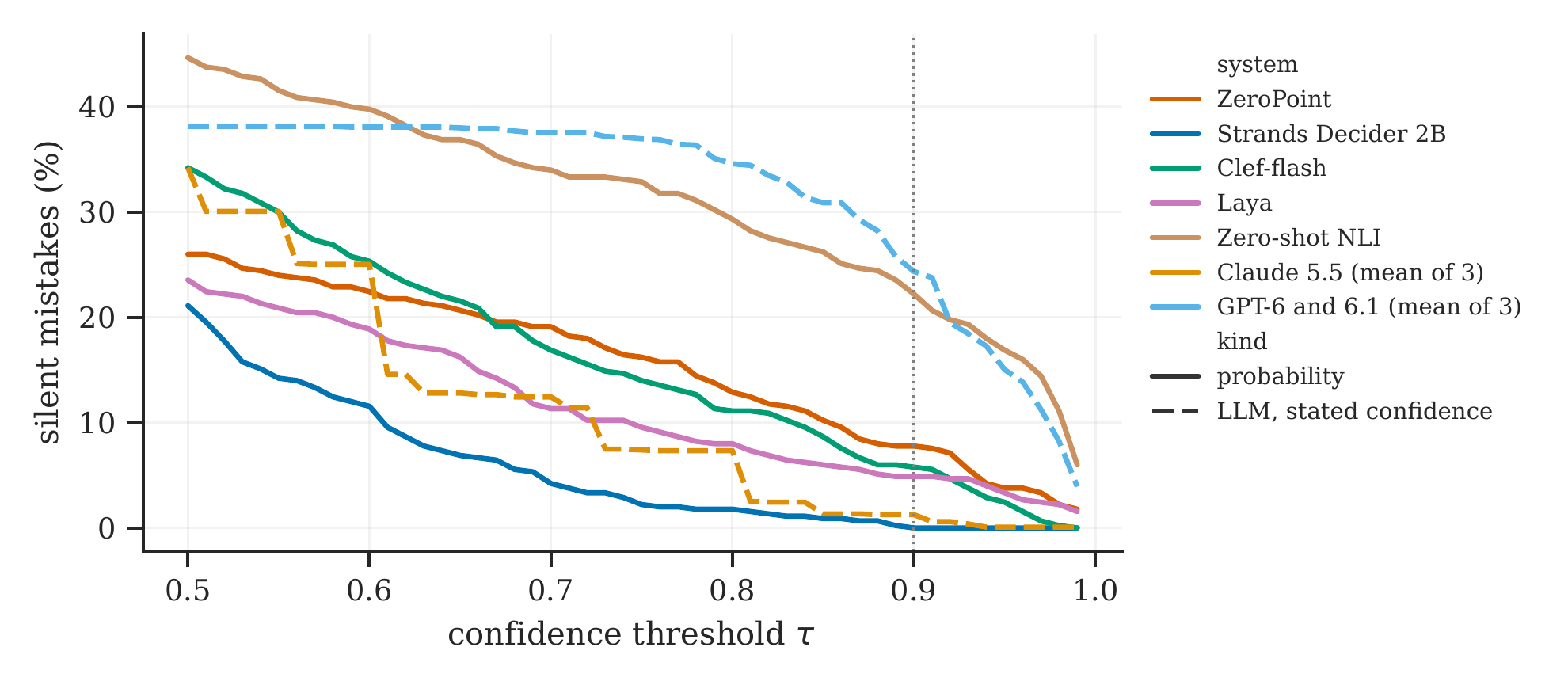}
\caption{Silent mistakes as a function of the confidence threshold $\tau$, on the 450 cases without a correct option. Solid lines use the probability returned by each system, and dashed lines the confidence stated by the hosted models, averaged within each family. The dotted line marks $\tau=0.9$, fixed before the results were seen.}
\label{fig:thresholds}
\end{figure}

\paragraph{Silent mistakes.} ZeroPoint makes silent mistakes in \zpSilent\% of the cases without a correct option, and Strands Decider in none. Most of ZeroPoint's occur on tickets whose own queue was removed, 18.7\%, where the remaining queues are close neighbours of the correct one. Strands Decider rarely states high confidence. It gives \stRightConf\% of its correct answers with probability 0.9 or more, against \zpRightConf\% for ZeroPoint. Figure~\ref{fig:thresholds} shows that Strands Decider makes the fewest silent mistakes at every threshold between 0.5 and 0.99, that ZeroPoint makes more than Laya at nearly every threshold, and that it makes more than Clef-flash above a threshold of 0.7. The GPT models state high confidence almost always and make 22.7\% to 26.0\% silent mistakes at the threshold of 0.9.

\paragraph{Order invariance, latency and size.} ZeroPoint's answer never changed with the order of the options. Strands Decider changed its answer in \stOrder\% of cases and Laya in \layaOrder\%, in line with the order sensitivity reported for language models~\cite{pezeshkpour2024large,zhao2021calibrate}. Clef-flash sorts the options and did not change. ZeroPoint takes \zpMs~ms per decision at the median on the 2-core CPU, with model files of 89~MB. Strands Decider weighs 4.8~GB and Clef-flash 19.1~GB. On the same 2-core CPU, over the first \sameCpuCases{} cases of the 2,000-case sample, ZeroPoint took \zpSameCpuMs~ms at the median and Strands Decider 6,770~ms, run in bfloat16 because float32 did not fit in memory.

\section{Discussion}
\label{sec:discussion}

Section~\ref{sec:gauge} shows that adding $c(X)$ to all energies leaves every comparison between options unchanged. Fixing the zero therefore cannot, by itself, improve the choice among options. It can only change comparisons between an energy and an absolute level, and the decision to abstain is such a comparison. Both ablations (Appendices~\ref{sec:lineage} and~\ref{sec:headablation}) agree that the fixed zero does not change the choice among options. They differ in how much it improves abstention. In the prototype, the fixed zero clearly outperformed a learned rejection class. In the released architecture, with only the output layer retrained, the advantage is small, and most of the benefit comes from training with examples in which no option fits. The argument of Section~\ref{sec:gauge} therefore identifies what a reference level can and cannot change, while the data show that a learned rejection class trained on the same examples recovers most of the same behaviour. The comparison with other systems shows the same division, although it does not isolate the cause. ZeroPoint ranks options worse than larger models and selects an option less often when none is correct. Ranking and abstention therefore appear to depend on different components. Ranking depends on the capacity of the encoder, and abstention on the reference level of the energy. A larger encoder could be combined with the same fixed zero without changing the argument.

Recent work trains an energy head to decide whether a retrieval-augmented language model should answer a medical question~\cite{shankar2025energy,shankar2025energysayno}. In both cases the energy is a learned score, and the published descriptions do not define a reference level for it. Classifiers can also be reinterpreted as joint energy-based models~\cite{grathwohl2019your}, and prediction with abstention has been studied as a single learning problem with formal guarantees~\cite{casacuberta2026selective}. Two other approaches place the decision to abstain elsewhere. Energy-based out-of-distribution detection~\cite{liu2020energy} reads the free energy of a classifier after training and thresholds it, which places the decision on the quantity shifted by the additive freedom. Selective classification~\cite{chow1970optimum,geifman2017selective} learns or calibrates a rejection rule beside the classifier. In ZeroPoint, the reference level is a state of the model, inside the distribution, trained by the same loss as the options.

The empty answer addresses a narrower question than out-of-distribution detection. It indicates that no listed option fits the text, not that the text lies far from the training data. A text unrelated to the task can still produce a shallow well for some option. In the prototype, a distance term on the encoder space detected such texts where the fixed zero alone did not, and the two signals answer different questions. ZeroPoint 1.0 includes only the first. The decline with list length (Appendix~\ref{sec:length}) shows a related limit of the same mechanism. A zero level prevents dilution by options of zero depth, not by many shallow ones.

\subsection{Limitations}
\label{sec:limits}

The additive freedom of Section~\ref{sec:gauge} is a property of the objective and holds by construction. Its empirical consequences are measured less completely. In the released architecture, the ablation retrains only the output layer on frozen features, so any effect of the fixed zero on the training of the encoder is not measured, and the output layers on the original encoder are weak. The released model was trained with one seed, and seed variability was measured on a preliminary version. The comparison with other systems rests on 900 cases per system, and another sample could move the results by a few points. The 2,000-case test was written with the help of a language model, and it is the test on which the larger systems lead most. The other decision models and the hosted models receive ``None of these'' as a text option, and a different wording could change how often they choose it. CPU, GPU and API latencies are not directly comparable. Texts foreign to the task were tested only with film reviews and newsgroup posts. ZeroPoint was trained on English and Spanish texts, and the other languages covered by its vocabulary were not evaluated one by one.

\section{Conclusion}

We have argued that energy-based decision models lack a reference level, because their training objective is invariant to adding any function of the input to all energies, and that this places the decision to abstain outside the model. ZeroPoint introduces that reference level as a state of the model with energy zero for every input, so that abstention becomes part of inference and is learned by the same likelihood as the choice among options. Ablations show that training with examples in which no option fits accounts for most of the benefit, and that the fixed zero adds a small but consistent improvement in separating inputs with and without a correct option. The resulting model is small enough to run on a CPU, is independent of the order of the options, and trades ranking accuracy for that efficiency. Future work should measure the effect of the fixed zero in the full architecture directly, and combine it with a larger encoder and with a distance term for inputs foreign to the task.

{\footnotesize\setlength{\bibsep}{1pt}
\bibliographystyle{unsrtnat}
\bibliography{references}}

@article{lecun2006tutorial,
  title={A tutorial on energy-based learning},
  author={LeCun, Yann and Chopra, Sumit and Hadsell, Raia and Ranzato, M and Huang, Fujie and others},
  journal={Predicting structured data},
  volume={1},
  number={0},
  year={2006}
}

@book{gibbs1902,
  title={Elementary principles in statistical mechanics: developed with especial reference to the rational foundations of thermodynamics},
  author={Gibbs, Josiah Willard},
  year={1902},
  publisher={C. Scribner's sons}
}

@article{chow1970optimum,
  title={On optimum recognition error and reject tradeoff},
  author={Chow, Chao},
  journal={IEEE Transactions on information theory},
  volume={16},
  number={1},
  pages={41--46},
  year={1970},
  publisher={IEEE}
}

@article{hendrycks2016baseline,
  title={A baseline for detecting misclassified and out-of-distribution examples in neural networks},
  author={Hendrycks, Dan and Gimpel, Kevin},
  journal={arXiv preprint arXiv:1610.02136},
  year={2016}
}

@article{liu2020energy,
  title={Energy-based out-of-distribution detection},
  author={Liu, Weitang and Wang, Xiaoyun and Owens, John and Li, Yixuan},
  journal={Advances in neural information processing systems},
  volume={33},
  pages={21464--21475},
  year={2020}
}

@article{geifman2017selective,
  title={Selective classification for deep neural networks},
  author={Geifman, Yonatan and El-Yaniv, Ran},
  journal={Advances in neural information processing systems},
  volume={30},
  year={2017}
}

@article{zaheer2017deep,
  title={Deep sets},
  author={Zaheer, Manzil and Kottur, Satwik and Ravanbakhsh, Siamak and Poczos, Barnabas and Salakhutdinov, Russ R and Smola, Alexander},
  journal={Advances in neural information processing systems},
  volume={30},
  year={2017}
}

@inproceedings{zhao2021calibrate,
  title={Calibrate before use: Improving few-shot performance of language models},
  author={Zhao, Zihao and Wallace, Eric and Feng, Shi and Klein, Dan and Singh, Sameer},
  booktitle={International conference on machine learning},
  pages={12697--12706},
  year={2021},
  organization={Pmlr}
}

@inproceedings{pezeshkpour2024large,
  title={Large language models sensitivity to the order of options in multiple-choice questions},
  author={Pezeshkpour, Pouya and Hruschka, Estevam},
  booktitle={Findings of the Association for Computational Linguistics: NAACL 2024},
  pages={2006--2017},
  year={2024}
}

@article{wang2024multilingual,
  title={Multilingual e5 text embeddings: A technical report},
  author={Wang, Liang and Yang, Nan and Huang, Xiaolong and Yang, Linjun and Majumder, Rangan and Wei, Furu},
  journal={arXiv preprint arXiv:2402.05672},
  year={2024}
}

@inproceedings{yin2019benchmarking,
  title={Benchmarking zero-shot text classification: Datasets, evaluation and entailment approach},
  author={Yin, Wenpeng and Hay, Jamaal and Roth, Dan},
  booktitle={Proceedings of the 2019 conference on empirical methods in natural language processing and the 9th international joint conference on natural language processing (EMNLP-IJCNLP)},
  pages={3914--3923},
  year={2019}
}

@article{he2021debertav3,
  title={Debertav3: Improving deberta using electra-style pre-training with gradient-disentangled embedding sharing},
  author={He, Pengcheng and Gao, Jianfeng and Chen, Weizhu},
  journal={arXiv preprint arXiv:2111.09543},
  year={2021}
}

@article{laurer2024less,
  title={Less annotating, more classifying: Addressing the data scarcity issue of supervised machine learning with deep transfer learning and BERT-NLI},
  author={Laurer, Moritz and Van Atteveldt, Wouter and Casas, Andreu and Welbers, Kasper},
  journal={Political Analysis},
  volume={32},
  number={1},
  pages={84--100},
  year={2024},
  publisher={Cambridge University Press}
}

@inproceedings{larson2019evaluation,
  title={An evaluation dataset for intent classification and out-of-scope prediction},
  author={Larson, Stefan and Mahendran, Anish and Peper, Joseph J and Clarke, Christopher and Lee, Andrew and Hill, Parker and Kummerfeld, Jonathan K and Leach, Kevin and Laurenzano, Michael A and Tang, Lingjia and others},
  booktitle={Proceedings of the 2019 Conference on Empirical Methods in Natural Language Processing and the 9th International Joint Conference on Natural Language Processing (EMNLP-IJCNLP)},
  pages={1311--1316},
  year={2019}
}

@inproceedings{casanueva2020efficient,
  title={Efficient intent detection with dual sentence encoders},
  author={Casanueva, I{\~n}igo and Tem{\v{c}}inas, Tadas and Gerz, Daniela and Henderson, Matthew and Vuli{\'c}, Ivan},
  booktitle={Proceedings of the 2nd workshop on natural language processing for conversational AI},
  pages={38--45},
  year={2020}
}

@inproceedings{fitzgerald2023massive,
  title={MASSIVE: A 1M-example multilingual natural language understanding dataset with 51 typologically-diverse languages},
  author={FitzGerald, Jack and Hench, Christopher and Peris, Charith and Mackie, Scott and Rottmann, Kay and Sanchez, Ana and Nash, Aaron and Urbach, Liam and Kakarala, Vishesh and Singh, Richa and others},
  booktitle={Proceedings of the 61st Annual Meeting of the Association for Computational Linguistics (Volume 1: Long Papers)},
  pages={4277--4302},
  year={2023}
}

@misc{bitext2023,
  author       = {{Bitext}},
  title        = {Customer Support, Retail Banking, Insurance and Telco {LLM} Chatbot Training Datasets},
  published = {Hugging Face Hub, CDLA-Sharing 1.0},
  url = "https://huggingface.co/datasets/bitext/Bitext-retail-banking-llm-chatbot-training-dataset",
  year         = {2023}
}

@article{coucke2018snips,
  title={Snips voice platform: an embedded spoken language understanding system for private-by-design voice interfaces},
  author={Coucke, Alice and Saade, Alaa and Ball, Adrien and Bluche, Th{\'e}odore and Caulier, Alexandre and Leroy, David and Doumouro, Cl{\'e}ment and Gisselbrecht, Thibault and Caltagirone, Francesco and Lavril, Thibaut and others},
  journal={arXiv preprint arXiv:1805.10190},
  year={2018}
}

@inproceedings{hemphill1990atis,
  title={The ATIS spoken language systems pilot corpus},
  author={Hemphill, Charles T and Godfrey, John J and Doddington, George R},
  booktitle={Speech and Natural Language: Proceedings of a Workshop Held at Hidden Valley, Pennsylvania, June 24-27, 1990},
  year={1990}
}

@inproceedings{li2021mtop,
  title={MTOP: A comprehensive multilingual task-oriented semantic parsing benchmark},
  author={Li, Haoran and Arora, Abhinav and Chen, Shuohui and Gupta, Anchit and Gupta, Sonal and Mehdad, Yashar},
  booktitle={Proceedings of the 16th Conference of the European Chapter of the Association for Computational Linguistics: Main Volume},
  pages={2950--2962},
  year={2021}
}

@inproceedings{liu2021benchmarking,
  title={Benchmarking natural language understanding services for building conversational agents},
  author={Liu, Xingkun and Eshghi, Arash and Swietojanski, Pawel and Rieser, Verena},
  booktitle={Increasing naturalness and flexibility in spoken dialogue interaction: 10th international workshop on spoken dialogue systems},
  pages={165--183},
  year={2021},
  organization={Springer}
}

@inproceedings{maas2011learning,
  title={Learning word vectors for sentiment analysis},
  author={Maas, Andrew and Daly, Raymond E and Pham, Peter T and Huang, Dan and Ng, Andrew Y and Potts, Christopher},
  booktitle={Proceedings of the 49th annual meeting of the association for computational linguistics: Human language technologies},
  pages={142--150},
  year={2011}
}

@incollection{lang1995newsweeder,
  title={Newsweeder: Learning to filter netnews},
  author={Lang, Ken},
  booktitle={Machine learning proceedings 1995},
  pages={331--339},
  year={1995},
  publisher={Elsevier}
}

@book{efron1994introduction,
  title={An introduction to the bootstrap},
  author={Efron, Bradley and Tibshirani, Robert J},
  year={1994},
  publisher={Chapman and Hall/CRC}
}

@inproceedings{koehn2004statistical,
  title={Statistical significance tests for machine translation evaluation},
  author={Koehn, Philipp},
  booktitle={Proceedings of the 2004 conference on empirical methods in natural language processing},
  pages={388--395},
  year={2004}
}

@misc{strands2026decider,
  author = {Strands Lab},
  title = {{Strands Decider}},
  year = {2026},
  howpublished = {Software, \texttt{strands-decider} 0.2.0, commit \texttt{216d1ccb}},
  url = {https://github.com/strands-labs/strands-decider},
  version = {0.3.0},
}

@misc{cloudflare2026clef,
  author = {{Cloudflare}},
  title = {Clef-flash},
  year = {2026},
  howpublished = {Open 9{B} decision model, commit \texttt{fb4417d3}},
  url = {https://developers.cloudflare.com/workers-ai/models/clef/},
}

@misc{convai2026laya,
  author = {NandakishorM, Conv AI},
  title = {{Laya}},
  year = {2026},
  howpublished = {Software, \texttt{laya} 0.4.21},
  url = {https://github.com/NandhaKishorM/laya},
  version = {0.4.21}, 
}

@misc{aurelio2024semanticrouter,
  author = {{Aurelio AI}},
  title = {Semantic Router},
  year = {2024},
  howpublished = {\url{https://github.com/aurelio-labs/semantic-router}, version 1.1.1},
  url = {https://github.com/aurelio-labs/semantic-router},
  version = {1.1.1},
}

@misc{anthropic2026claude,
  author       = {{Anthropic}},
  title        = {{Claude}},
  year         = {2026},
  howpublished = {Large language model},
  url          = {https://claude.ai},
  note         = {Accessed: 2026-10-11}
}

@misc{openai2026gpt6,
  author       = {{OpenAI}},
  title        = {{GPT 6}},
  year         = {2026},
  howpublished = {Large language model},
  url          = {https://chatgpt.com},
  note         = {Accessed: 2026-10-11}
}

@misc{onnxruntime,
  author = {{ONNX Runtime developers}},
  title = {{ONNX} Runtime},
  year = {2021},
  howpublished = {\url{https://onnxruntime.ai}},
}

@misc{marin2026zeropoint,
  author = {Mar{\'i}n, Javier},
  title = {{ZeroPoint} 1.1.0},
  year = {2026},
  howpublished = {\url{https://github.com/Javihaus/zeropoint}},
  version = {1.0.0},
}

@article{yalovetzky2026lehami,
  title={LeHami: Symplectic Latent Dynamics for World Models},
  author={Yalovetzky, Romina and Balestriero, Randall and LeCun, Yann},
  journal={arXiv preprint arXiv:2610.15267},
  year={2026}
}

@article{greydanus2019hamiltonian,
  title={Hamiltonian neural networks},
  author={Greydanus, Samuel and Dzamba, Misko and Yosinski, Jason},
  journal={Advances in neural information processing systems},
  volume={32},
  year={2019}
}

@article{schick2023toolformer,
  title={Toolformer: Language models can teach themselves to use tools},
  author={Schick, Timo and Dwivedi-Yu, Jane and Dess{\`\i}, Roberto and Raileanu, Roberta and Lomeli, Maria and Hambro, Eric and Zettlemoyer, Luke and Cancedda, Nicola and Scialom, Thomas},
  journal={Advances in neural information processing systems},
  volume={36},
  pages={68539--68551},
  year={2023}
}

@article{shankar2025energy,
  title={Energy Landscapes Enable Reliable Abstention in Retrieval-Augmented Large Language Models for Healthcare},
  author={Shankar, Ravi and Wong, Sheng and Li, Lin and Bachmann, Magdalena and Silverthorne, Alex and Albert, Beth and Jones, Gabriel Davis},
  journal={arXiv preprint arXiv:2509.04482},
  year={2025}
}

@inproceedings{shankar2025energysayno,
  title={The Energy to Say No: Pre-Generation Abstention for Safety-Critical Medical RAG},
  author={Shankar, Ravi and Wong, Sheng Fung and Li, Lin and Bachmann, Magdalena and Silverthorne, Alex and Albert, Beth and Jones, Gabriel Davis},
  booktitle={The Second Workshop on GenAI for Health: Potential, Trust, and Policy Compliance},
  year= {2025}
}

@article{grathwohl2019your,
  title={Your classifier is secretly an energy based model and you should treat it like one},
  author={Grathwohl, Will and Wang, Kuan-Chieh and Jacobsen, J{\"o}rn-Henrik and Duvenaud, David and Norouzi, Mohammad and Swersky, Kevin},
  journal={arXiv preprint arXiv:1912.03263},
  year={2019}
}

@article{casacuberta2026selective,
  title={Selective omniprediction and fair abstention},
  author={Casacuberta, S{\'\i}lvia and Kanade, Varun},
  journal={Advances in Neural Information Processing Systems},
  volume={38},
  pages={39435--39477},
  year={2026}
}

@article{holm1979simple,
  title={A simple sequentially rejective multiple test procedure},
  author={Holm, Sture},
  journal={Scandinavian journal of statistics},
  pages={65--70},
  year={1979},
  publisher={JSTOR}
}

\appendix

\section{Ablation in a prototype}
\label{sec:lineage}

The argument of Section~\ref{sec:gauge} predicts that a model trained with a fixed zero should separate inputs with and without an acceptable option better than a model that learns rejection as one more class, and that it should not choose better among the options. We tested this before building ZeroPoint 1.0, with a prototype that shares its principle but not its architecture. The prototype uses a frozen e5-base encoder and places the wells on its normalized embeddings. It was compared with two controls trained on the same data, with the same encoder and the same examples without a correct option. The first control is a softmax over the options, and the second adds a learned ``none of the above'' class. The hypotheses and tests were fixed before training, and differences are reported with paired bootstrap intervals and Holm correction for multiple comparisons. Table~\ref{tab:ablation} gives the results. On support tickets whose own queue was removed from the list, the fixed zero separates the two kinds of ticket better than the learned class, by $+0.081$ (95\% interval $[0.049,0.117]$, Holm-corrected $p=0.0006$), with a standard deviation of 0.008 over five seeds. On queues never removed during training, the learned class falls to chance while the fixed zero does not, a difference of $+0.20$ ($[0.11,0.29]$). On out-of-scope questions written by people, the fixed zero also exceeds the plain softmax, by $+0.024$ ($[0.017,0.031]$, Holm-corrected $p=0.004$). As predicted, the choice among options does not improve, and accuracy stays within 3 points of the softmax on each of eight datasets. The prototype has two limitations. It does not detect texts entirely foreign to the task, because normalized embeddings place every text near some well, and with an argmax rule it abstained on 69\% of the tickets that had a correct queue. ZeroPoint 1.0 keeps the fixed zero, reads the text and all options together, and fine-tunes the encoder. Appendix~\ref{sec:headablation} measures the effect in that architecture.

\begin{table}[t]
\centering
\small
\begin{tabular}{lcccc}
\toprule
Test (AUROC) & Fixed zero & Softmax & Softmax + none & Other \\
\midrule
Tickets, queue removed, within each queue & \textbf{0.679} & 0.556 & 0.599 & 0.615$^{a}$ \\
\quad the same, queues never removed in training & \textbf{0.672} & -- & 0.472 & -- \\
Tickets, queue removed, second experiment & \textbf{0.730} & 0.514 & -- & 0.488$^{b}$ \\
Out-of-scope questions written by people & \textbf{0.987} & 0.963 & -- & 0.962$^{b}$ \\
Foreign texts (film reviews, newsgroup posts) & 0.612 & 0.596 & -- & \textbf{1.000}$^{b}$ \\
\bottomrule
\end{tabular}
\caption{Ablation. The prototype with the empty answer fixed at zero against softmax controls trained on the same data with the same frozen encoder. AUROC for separating inputs with and without a correct option. The first two rows come from the first experiment (fixed zero over 5 seeds, $0.670\pm0.008$) and the others from the second (2 seeds). $^{a}$Energy of Laya~\cite{convai2026laya}. $^{b}$Mahalanobis distance. Tests built from support tickets, CLINC150~\cite{larson2019evaluation}, film reviews~\cite{maas2011learning} and newsgroup posts~\cite{lang1995newsweeder}.}
\label{tab:ablation}
\end{table}

\section{Ablation of the output distribution}
\label{sec:headablation}

To measure the effect in the released architecture without retraining the full model, we froze the encoder and trained only the output layer, so that every variant shares the same features and data. We used two encoders. The released encoder was fine-tuned with the fixed zero and may favour it. The original multilingual-E5-small~\cite{wang2024multilingual} was never trained with any of the variants. For each encoder we trained five output layers with five seeds each, on 30,000 of the 76,211 training lists with the same mix of list types. The fixed zero uses \eqref{eq:weights}. The constant empty weight uses $e^{a/T}$ for the options and $e^b$ for the empty answer, which isolates the subtraction of one. The learned rejection class gives the empty answer a weight $e^{r(X)}$ computed from the text. The softmax is trained only on lists with a correct option and abstains below a probability threshold fitted on held-out lists. The last variant replaces the softplus with a ReLU, so that an exact zero depth is reachable. The protocol and the comparisons were fixed before training. Tables~\ref{tab:headablation} and~\ref{tab:headpaired} give the results.

Three results stand out. First, training with examples in which no option fits accounts for most of the abstention behaviour. On the released encoder, the softmax trained without them makes 8.6 points fewer correct decisions than the fixed zero and has an FPR95 36.5 points higher, while the learned rejection class reaches the same rate of correct decisions as the fixed zero on both encoders. Second, the fixed zero separates inputs with and without a correct option slightly better than the learned rejection class on both encoders, by 0.002 and 0.003 in AUROC, and lowers FPR95 by 2.0 points on the released encoder, whereas the learned rejection class makes fewer silent mistakes and has a lower calibration error. Third, the subtraction of one in \eqref{eq:weights} improves separation only on the released encoder, which was trained with it, where it lowers FPR95 by 15.6 points. On the original encoder the two forms differ by less than 0.001 in AUROC and not at all in correct decisions or FPR95. Exact zero depths do not help. With ReLU depths, correct decisions are unchanged on the released encoder, but separation is worse and unstable across seeds, and on the original encoder every measure is worse. The output layers trained on the original encoder reach far lower accuracy than those on the released one, so their comparisons rest on weak features. This design does not measure any effect of the fixed zero on the training of the encoder itself.

\begin{table}[t]\centering\small\resizebox{\linewidth}{!}{\begin{tabular}{llrrrrrrr}\toprule
Encoder & Output layer & Correct & Ranking & AUROC & FPR95 & Silent & False & ECE \\ & & decisions & & & & mistakes & rejections & \\\midrule
Released & Fixed zero, $e^{a/T}-1$ & 69.7 & 76.0 & 0.861 & 48.3 & 5.2 & 23.8 & 15.4 \\
 & Constant empty weight, $e^{a/T}$ & 69.2 & 75.3 & 0.850 & 63.5 & 5.8 & 23.4 & 16.5 \\
 & Learned rejection class & 69.6 & 75.9 & 0.859 & 50.3 & 3.9 & 21.7 & 15.1 \\
 & Softmax with threshold & 61.3 & 75.9 & 0.711 & 84.9 & 28.9 & 29.8 & 27.1 \\
 & Fixed zero, ReLU depths & 69.3 & 72.3 & 0.850 & 81.0 & 3.8 & 19.8 & 15.6 \\
\midrule
Original & Fixed zero, $e^{a/T}-1$ & 44.7 & 45.2 & 0.676 & 83.0 & 1.8 & 34.0 & 9.7 \\
 & Constant empty weight, $e^{a/T}$ & 44.7 & 45.1 & 0.675 & 83.1 & 1.7 & 34.0 & 9.9 \\
 & Learned rejection class & 43.8 & 45.7 & 0.672 & 82.5 & 0.8 & 29.7 & 7.4 \\
 & Softmax with threshold & 38.9 & 46.5 & 0.587 & 91.7 & 5.7 & 32.3 & 31.2 \\
 & Fixed zero, ReLU depths & 32.6 & 31.3 & 0.571 & 92.7 & 0.0 & 37.3 & 5.4 \\
\bottomrule\end{tabular}}
\caption{Output-layer ablation on the eight tests, 7,537 cases, mean over 5 seeds, in percent except AUROC. Only the output layer is trained, on frozen features from the released encoder or from the original multilingual-E5-small~\cite{wang2024multilingual}, with the same 30,000 training lists for every variant. Standard deviations over seeds are below 0.3 points for every variant except the ReLU depths on the released encoder (FPR95 standard deviation 26 points). False rejections are abstentions on cases with a correct option.}\label{tab:headablation}\end{table}

\begin{table}[t]\centering\small\resizebox{\linewidth}{!}{\begin{tabular}{llccc}\toprule
Encoder & Fixed zero minus & Correct decisions & AUROC & FPR95 \\\midrule
Released & Constant empty weight, $e^{a/T}$ & +0.4 [+0.1, +0.7]$^{*}$ & +0.011 [+0.010, +0.013]$^{*}$ & $-$15.6 [$-$17.2, $-$13.8]$^{*}$ \\
 & Learned rejection class & +0.1 [$-$0.2, +0.4] & +0.002 [+0.001, +0.002]$^{*}$ & $-$2.0 [$-$3.0, $-$0.8]$^{*}$ \\
 & Softmax with threshold & +8.6 [+7.5, +9.7]$^{*}$ & +0.150 [+0.138, +0.162]$^{*}$ & $-$36.5 [$-$39.3, $-$33.7]$^{*}$ \\
 & Fixed zero, ReLU depths & +0.3 [$-$0.1, +0.7] & +0.005 [+0.004, +0.005]$^{*}$ & $-$4.4 [$-$6.2, $-$2.8]$^{*}$ \\
\midrule
Original & Constant empty weight, $e^{a/T}$ & +0.1 [$-$0.1, +0.2] & +0.001 [+0.000, +0.001]$^{*}$ & $-$0.2 [$-$0.6, +0.4] \\
 & Learned rejection class & +0.6 [+0.1, +1.2] & +0.003 [+0.001, +0.005]$^{*}$ & +0.5 [$-$0.7, +1.6] \\
 & Softmax with threshold & +6.0 [+4.8, +7.2]$^{*}$ & +0.088 [+0.072, +0.104]$^{*}$ & $-$8.7 [$-$10.7, $-$6.1]$^{*}$ \\
 & Fixed zero, ReLU depths & +10.1 [+9.0, +11.3]$^{*}$ & +0.099 [+0.088, +0.110]$^{*}$ & $-$9.8 [$-$11.4, $-$7.9]$^{*}$ \\
\bottomrule\end{tabular}}
\caption{Paired differences between the fixed zero and each other output layer, on the same 7,537 cases, with per-case probabilities averaged over seeds and 95\% bootstrap intervals over 2,000 resamples stratified by test~\cite{efron1994introduction}. Correct decisions and FPR95 in percentage points. $^{*}$Significant after Holm correction over the 24 comparisons~\cite{holm1979simple}. For FPR95, a negative value favours the fixed zero.}\label{tab:headpaired}\end{table}

\section{Eight tests on a CPU}
\label{sec:eight}

Table~\ref{tab:eight} adds five public tests for the systems that run on a CPU. The larger models are not evaluated on them, since they may have seen them in training. Among these four systems, ZeroPoint has the highest rate of correct decisions, the lowest FPR95, the highest AUROC and the fewest silent mistakes, while Laya ranks the correct option first more often. ZeroPoint was trained on the training split of CLINC150, so that test lies in a domain it has seen.

\begin{table}[t]
\centering
\small
\begin{tabular}{lrrrrrrr}
\toprule
System & Correct & Ranking & Silent & FPR95 & AUROC & Order & Latency \\
 & decisions & & mistakes & & & changes & (median) \\
\midrule
\textbf{ZeroPoint} & 63.6 & 68.5 & 6.9 & 53.6 & 0.808 & 0.0 & 21 ms \\
Laya & 59.8 & 72.2 & 21.4 & 60.1 & 0.794 & 14.9 & 383 ms \\
Semantic Router & 34.6 & 56.8 & -- & 69.0 & 0.740 & 0.2 & 9 ms \\
Zero-shot NLI & 28.6 & 50.6 & 30.2 & 87.1 & 0.666 & 0.0 & 1,316 ms \\
\bottomrule
\end{tabular}
\caption{Mean over eight tests, 7,537 cases, for the systems that run on a CPU, in percent except AUROC and latency. AUROC measures the separation of cases with and without a correct option~\cite{hendrycks2016baseline}. FPR95 is the share of cases without a correct option that are still acted on when the threshold keeps 95\% of the cases with one. The tests are the three of Table~\ref{tab:main}, SNIPS~\cite{coucke2018snips}, ATIS~\cite{hemphill1990atis}, MTOP in Spanish~\cite{li2021mtop}, HWU64~\cite{liu2021benchmarking} and CLINC150 with 20 options~\cite{larson2019evaluation}. Zero-shot NLI reads the 300-case sample of each test.}
\label{tab:eight}
\end{table}

\section{Length of the list}
\label{sec:length}

Property 1 of Section~\ref{sec:model} holds exactly for options of zero depth and approximately for shallow ones. Table~\ref{tab:length} measures how the released model behaves as the list grows, with lists of 20 to 150 options from CLINC150 and of 20 and 50 options from MTOP in Spanish, 600 cases each, half of them without a correct option. On CLINC150, correct decisions fall from 90.5\% with 20 options to 68.2\% with 150, and silent mistakes rise from 0.7\% to 7.0\%. Shallow but non-zero wells therefore accumulate weight as the list grows, as the condition $\sum_y a_y/T\ll e^b$ in property 1 anticipates. Lists longer than 20 options are also read in separate blocks whose options do not attend to each other. The decline is steeper on MTOP in Spanish, where ranking is already weaker with 20 options. The residual weight of the options in cases without a correct option grows with the list. Relative to the weight $e^b$ of the empty answer, its median rises from 0.07 with 20 options to 2.07 with 150 on CLINC150, and the share of such cases in which the options outweigh the empty answer rises from 6\% to 74\%.

\begin{table}[t]
\centering
\small
\begin{tabular}{lrrrrr}
\toprule
Test & Options & Correct decisions & Ranking & Silent mistakes & AUROC \\
\midrule
CLINC150 & 20 & 90.5 & 95.3 & 0.7 & 0.973 \\
 & 50 & 83.5 & 92.0 & 6.0 & 0.934 \\
 & 100 & 74.0 & 85.3 & 4.7 & 0.915 \\
 & 150 & 68.2 & 83.0 & 7.0 & 0.898 \\
\midrule
MTOP (Spanish) & 20 & 53.3 & 60.0 & 4.7 & 0.659 \\
 & 50 & 34.7 & 42.7 & 13.3 & 0.564 \\
\bottomrule
\end{tabular}
\caption{The released model files on lists of increasing length, 600 cases per row, half of them without a correct option, in percent except AUROC. Lists built from CLINC150~\cite{larson2019evaluation} and MTOP~\cite{li2021mtop}.}
\label{tab:length}
\end{table}

\section{Adaptation to new labels}
\label{sec:fit}

The released model can be fine-tuned on a small set of labelled texts, with lists built from those labels and 30\% of them without a correct option, for a fixed 200 steps. Table~\ref{tab:fit} reports the two ticket tests, 600 cases each, before and after adaptation with 100 to 1,000 labelled tickets kept separate from the test cases. With all layers trained, ranking on tickets whose queue was removed rises from 24.3\% to between 36.0\% and 39.7\%, and silent mistakes fall from 19.3\% to between 6.7\% and 0\%. Keeping the first 8 encoder layers fixed gives similar ranking and no silent mistakes, and it preserves the separation of unrelated texts, with an AUROC of 0.991 against 0.952 when all layers are trained. On a 2-core CPU, an adaptation takes 11 minutes with 8 layers fixed and 18 minutes with all layers trained.

\begin{table}[t]
\centering
\small
\begin{tabular}{llrrrrrr}
\toprule
 & & \multicolumn{3}{c}{Tickets, queue removed} & \multicolumn{3}{c}{Tickets, unrelated texts} \\
\cmidrule(lr){3-5}\cmidrule(lr){6-8}
Layers trained & Examples & Ranking & Silent & AUROC & Ranking & Silent & AUROC \\
\midrule
None (released) & -- & 24.3 & 19.3 & 0.508 & 28.7 & 0.3 & 0.993 \\
\midrule
All & 100 & 36.0 & 6.7 & 0.652 & 44.7 & 2.7 & 0.918 \\
All & 300 & 38.0 & 2.0 & 0.692 & 42.0 & 1.7 & 0.952 \\
All & 1,000 & 39.7 & 0.0 & 0.705 & 46.7 & 0.3 & 0.947 \\
\midrule
Last 4 of 12 & 300 & 36.0 & 0.0 & 0.594 & 46.7 & 0.0 & 0.991 \\
Last 4 of 12 & 1,000 & 36.7 & 0.0 & 0.584 & 46.3 & 0.0 & 0.993 \\
\bottomrule
\end{tabular}
\caption{Adaptation of the released model with 200 optimizer steps on labelled tickets kept separate from the test cases. Ranking and silent mistakes in percent. Each test has 600 cases, half of them without a correct option.}
\label{tab:fit}
\end{table}

\section{Variability and reproducibility}
\label{sec:repro}

Table~\ref{tab:repro} summarizes the sources of variation measured on the model. Two training seeds of a preliminary version with the same architecture, trained before the vocabulary reduction, the hard negatives, the long lists and the refit of $T$ and $b$ were added, differed on average over the eight tests by 1.4 points in ranking and by 0.005 in AUROC, and by at most 3.4 points in ranking or correct decisions on any test. For the training checkpoint, five orders of the options on 300 cases of each of five tests produced identical probabilities. For the training checkpoint, probabilities computed with 1, 2 and 4 CPU threads and on a GPU or a CPU differ by less than $3.2\times10^{-6}$. The released files store the weights in float16 and the vocabulary embeddings in int8. Against the checkpoint, this storage changes 11 of 7,537 decisions on the eight tests, with a largest probability difference of 0.038. The released files give probabilities that agree within $2.9\times10^{-6}$ on Linux (x86 and ARM), macOS and Windows with Python 3.10, 3.13 and 3.14, and no decision changes.

\begin{table}[t]
\centering
\small
\begin{tabular}{llll}
\toprule
Source of variation & Model & Measurement & Result \\
\midrule
Training seed (2 seeds) & preliminary & mean $|\Delta|$, 8 tests & 1.4 pts, 0.005 AUROC \\
Order (5 orders, 1,500 cases) & checkpoint & max $|\Delta p|$ & 0 \\
CPU threads (1, 2 and 4) & checkpoint & max $|\Delta p|$ & $2.2\times10^{-6}$ \\
GPU against CPU & checkpoint & max $|\Delta p|$ & $3.2\times10^{-6}$ \\
Float16 and int8 storage & released & decisions changed & 11 of 7,537 \\
Operating system, Python & released & max $|\Delta p|$ & $2.9\times10^{-6}$ \\
\bottomrule
\end{tabular}
\caption{Variability of the released model. $\Delta p$ is the difference in any probability between two runs on the same cases. Apart from the training seed and the storage format, no source of variation changed any decision.}
\label{tab:repro}
\end{table}

\end{document}